\pdfoutput=1

\documentclass[11pt]{article}

\usepackage[]{acl}

\usepackage{times}
\usepackage{latexsym}

\usepackage[T1]{fontenc}

\usepackage[utf8]{inputenc}

\usepackage{microtype}
\usepackage{enumitem}
\usepackage{algorithm}
\usepackage{algorithmic}
\usepackage{xcolor}
\usepackage{graphicx} 
\usepackage{amsmath}
\usepackage{booktabs}
\usepackage{multirow}
\usepackage{arydshln}
\usepackage{array}
\usepackage{amssymb}
\usepackage{bm}
\usepackage{bbding}
\usepackage{color}
\usepackage{siunitx}
\usepackage{makecell}
\usepackage{multirow}
\definecolor{c1}{RGB}{255, 204, 210}
\definecolor{c2}{RGB}{255, 213, 158}
\definecolor{c3}{RGB}{161, 227, 216}
\definecolor{c4}{RGB}{154, 220, 255}

\title{Tailor: A Prompt-Based Approach to Attribute-Based \\  Controlled Text Generation}

\author{Kexin Yang$^\spadesuit{\,}$\thanks{\hspace{2mm}Work is done during internship at DAMO Academy} \:\:\: \textbf{Dayiheng Liu}$^\spadesuit{\,}$ \:\:\: \textbf{Wenqiang Lei}$^\diamondsuit$ \:\:\: \textbf{Baosong Yang}$^\spadesuit{\,}$ \:\:\: \textbf{Mingfeng Xue}\\ \:\:\:\textbf{Boxing Chen}$^{\spadesuit}$ \:\:\:\textbf{Jun Xie}$^{\spadesuit}$ \:\:\:\\
$^\spadesuit$Alibaba Group \\
$^\diamondsuit$National University of Singapore \\ 
\texttt{\{kexinyang0528, losinuris\}@gmail.com}
}

\begin{document}
\maketitle
\begin{abstract}

Attribute-based Controlled Text Generation (CTG) refers to generating sentences that satisfy desirable attributes (e.g., emotions and topics). Existing works often utilize fine-tuning or resort to extra attribute classifiers, yet suffer from storage and inference time increases. To address these concerns, we explore attribute-based CTG in a prompt-based manner. In short, the proposed Tailor represents each attribute as a pre-trained continuous vector (i.e., single-attribute prompt) and guides the generation of a fixed PLM switch to a pre-specified attribute. We experimentally find that these prompts can be simply concatenated as a whole to multi-attribute CTG without any re-training, yet raises problems of fluency decrease and position sensitivity. To this end, Tailor provides a multi-attribute prompt mask and a re-indexing position-ids sequence to bridge the gap between the training (one prompt for each task) and testing stage (concatenating more than one prompt). To further enhance such single-attribute prompt combinations, Tailor also introduces a trainable prompt \textit{connector}, which can be concatenated with any two single-attribute prompts to multi-attribute text generation. Experiments on 11 attribute-specific generation tasks demonstrate strong performances of Tailor on both single-attribute and multi-attribute CTG, with 0.08\% training parameters of a GPT-2.



\end{abstract}

\section{Introduction}
%

%

Attribute-based CTG~\cite{ctg_survey} focuses on generating sentences satisfying pre-specified attributes such as topic and sentiment, which remains extremely challenging in recent progress~\cite{pplm}. Especially multi-attribute CTG, it is typically unsupervised since no example of a sentence with specified attributes could be obtained during training.~\cite{multiple_attribute_text}. Existing efforts for attribute-based CTG can be roughly divided into two types: fine-tuning and utilizing extra attribute classifiers. The first type usually fine-tunes a pre-trained language model (PLM) on the attribute-specific data~\cite{finetune_attribute}, yet stores a full copy of the PLM for each desirable attribute. To partly address this issue, control codes are introduced to generate various styles of sentences with one PLM, such as keywords~\cite{ctrl} and numerical sequence~\cite{style_ptb}. However, re-training whole PLMs could be expensive~\cite{fudge} and they rarely attend to multi-attribute CTG. The second type introduces extra attribute classifiers to guide a PLM, such as back-propagating gradients of classifiers~\cite{pplm} or weighting output logits~\cite{gedi,fudge}. Such a paradigm shows encourage improvement, while the text fluency tends to decrease (see \S~\ref{sec:main_results}) and inference time increase~\cite{control_prefix}. 

\begin{figure*}[t]
  \centering
  \includegraphics[scale=0.38]{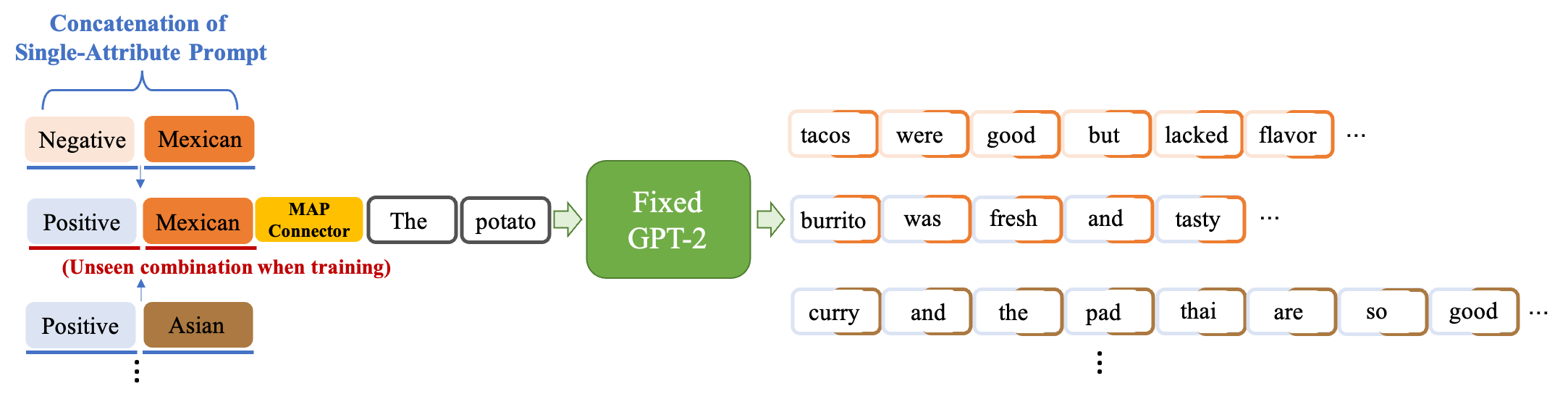}
  \caption{An overview of MAP connector to multi-attribute CTG. MAP connector concatenates single-attribute prompts (i.e., underlines), then continues with a pre-specified input prefix to the fixed GPT-2. Note that, it is able to the unseen combination (e.g, sentiment of Positive and topic of Mexican food).}
  \label{fig:figure1}
\end{figure*}

To overcome the aforementioned limitations, we propose a \textbf{T}ext-\textbf{a}ttr\textbf{i}bute control\textbf{lor} (\textbf{Tailor}) --- a prompt-based approach to attribute-based CTG. The key idea is to represent each attribute as a pre-trained continuous vector (hereinafter known as the single-attribute prompt) to control a fixed GPT-2 for single-attribute CTG, and effectively concatenate such single-attribute prompts as a whole for multi-attribute CTG. This allows Tailor to be easily expanded by training the corresponding attribute prompt if a new attribute emerges, while avoiding re-training the whole PLM. In detail, the single-attribute prompt is concatenated with the input prefix and then guides the generation of a fixed GPT-2 switch to a pre-specified attribute. More importantly, we experimentally find that such single-attribute prompts could be simply concatenated to generate sentences with multi attributes. However, this manner always suffers from fluency decrease and position sensitivity, i.e., the PLM tends to focus more on the single-attribute prompt that is closer to the input prefix (see \S~\ref{sec:experiment_discussions}). To address these issues, the key lies in bridging the gap between the training and the testing stage. In detail, the single-attribute prompt only attends to itself in the attention matrix while training, since it is individually trained by the attribute-specific data. However, when it comes to the testing stage, the second prompt also attends to the first one in the concatenation, with the simultaneous change of the position-ids sequence\footnote{In this case, position-ids sequence denotes position indexes of input tokens in the position embeddings for GPT-2.}.

To fill this gap, Tailor introduces a \textbf{M}ulti-\textbf{A}ttribute \textbf{P}rompt mask (\textbf{MAP} mask) and a \textbf{R}e-indexing \textbf{P}osition-ids sequence (\textbf{RP} sequence) for the fixed GPT-2. MAP mask prevents distinct single-attribute prompts from cross-attention, and RP sequence ensures stable position-ids information for the PLM after swapping, by individually numbering each prompt. As such non-training method partly addresses the issue, the text fluency still decrease, since there is no multi-attribute specific training stage for these single-attribute prompts to adapt to work together.  Inspired by the role of `and' in connecting parallel phrases for natural sentences~\cite{androle}, Tailor further provides a training method that contains a continuous connector to connect two single-attribute prompts as a whole to multi-attribute CTG.  As shown in Figure~\ref{fig:figure1}, the proposed \textbf{M}ulti-\textbf{A}ttribute \textbf{P}rompt connector (\textbf{MAP} connector) can be concatenated with any two singe-attribute prompts and hint a GPT-2 to multi-attribute CTG. Meanwhile, a pseudo-prompt based strategy is also provided for training the connector in unsupervised settings. With MAP connector, the combinations show strong performances on multi-attribute CTG on the popular benchmark YELP dataset~\cite{multiple_attribute_text}. Furthermore, MAP connector can get encouraging improvements for the combinations even if they are unseen in the connector's training stage. The main contributions are as follows:

\begin{itemize}
    \item We propose \textbf{Tailor}, a prompt-based approach to attribute-based CTG. To jointly include both single-attribute and multi-attribute CTG in an unified paradigm, Tailor employs a set of pre-trained continuous prefixes to guide a fixed PLM to switch to a pre-specified attribute, and effectively concatenating them to generate multi-attribute sentences.
    
    \item We experimentally reveal the combining ability of continuous prompts. To enhance this combination, we explore two effective strategies without training (MAP mask + RP sequence) or with training (MAP connector) after single-attribute CTG. Especially, MAP connector achieves strong performances on six multi-attribute generation tasks, even works to the unseen ones. 
\end{itemize}

\section{Related Work}
\noindent\textbf{Attribute-Based CTG} focuses on generating sentences containing pre-specified attributes, such as sentiment and topic. As a vital demand for intelligent writing~\cite{ctg_survey}, various attempts have been made in this area, including fine-tuning PLMs and utilizing extra attribute classifiers. 
The first type usually fine-tunes separately and stores a full copy of PLM for each desirable attribute ~\cite{finetune_attribute}. To alleviate the storage problem, CTRL~\cite{ctrl} provides 55 kinds of control codes (i.e., special keywords) to fine-tune one PLM for generating sentences of various styles. StylePTB~\cite{style_ptb} also proposes several style transfer tokens (i.e., a sequence of numbers) to guide a GPT-2~\cite{gpt2} to multiple styles transfer. GSum~\cite{gsum} introduces four guidance signals (e.g., keywords and relations) to enhance the controllability of PLMs in the text summarization. Although they make successful attempts in single-attribute CTG and partially address the storage issue~\cite{fudge}, it might not be directly usable for multi-attribute CTG. To improve the flexibility and extensibility of the CTG model, the second type makes efforts in the inference stage. In short, utilizing extra attribute classifiers to guide PLMs in each generating step. PPLM~\cite{pplm} iteratively modifies latent representations of a GPT-2 referring to the gradient of attribute classifiers, yet notably increasing the inference time. To solve this problem, Fudge~\cite{fudge} uses an attribute predictor to adjust the output probabilities of a PLM. Similarly, GeDi~\cite{gedi} uses smaller PLMs as generative discriminators to hint a larger PLM generating sentences that satisfy desirable attributes. Despite their progress, the fluency of generating sentences tends to decrease compared with the original PLM (see \S~\ref{sec:main_results}) and extra inference time costs still existed. In comparison, with Tailor, PLMs can enjoy the benefits of the controllability from combinations of single-attribute prompts with a negligible decrease of text quality. \par
\noindent\textbf{Prompt Learning} is a new paradigm in NLP summarised as ``Pre-train, Prompt and Predict''~\cite{prompt_survey}. In short, it can guide a single PLM to solve various downstream tasks by reformulating these tasks into a text-to-text manner. Early works explore prompt formatting as discrete-word templates~\cite{lama,pet,autoprompt,adapter}. Recently, continuous prompt has attracted attention~\cite{ppt,p_tuning_v1,p_tuning_v2}, which usually forms as a set of continuous task-specific vectors to the input. Unlike discrete prompts may face difficulties of optimizing, continuous prompts could be trained expressively on downstream task data~\cite{prefix-tuning}. Despite their encouraging progress, the prompt composition is rarely explored but undoubtedly important in prompt learning. In that case, a composable task could be accomplished by composing various subtasks with multiple sub-prompts~\cite{prompt_survey}. To achieve it, PTR~\cite{ptr} introduces manual sub-prompts for entity recognition and relation classification, respectively. Then, these two kinds of prompts are composed by logic rules as a complete prompt for the relation extraction task. Unfortunately, the composition of continuous prompts is rarely explored yet has demonstrated great potential~\cite{control_prefix}. In this paper, we experimentally reveal the potential of combining continuous prompts to accomplish multi-attribute CTG. Afterward, we propose MAP connector to enhance this combination. Extensive experiments verify the effectiveness of Tailor on both controllability of attributes and text quality.


\begin{figure*}[h]
  \centering
  \includegraphics[scale=0.35]{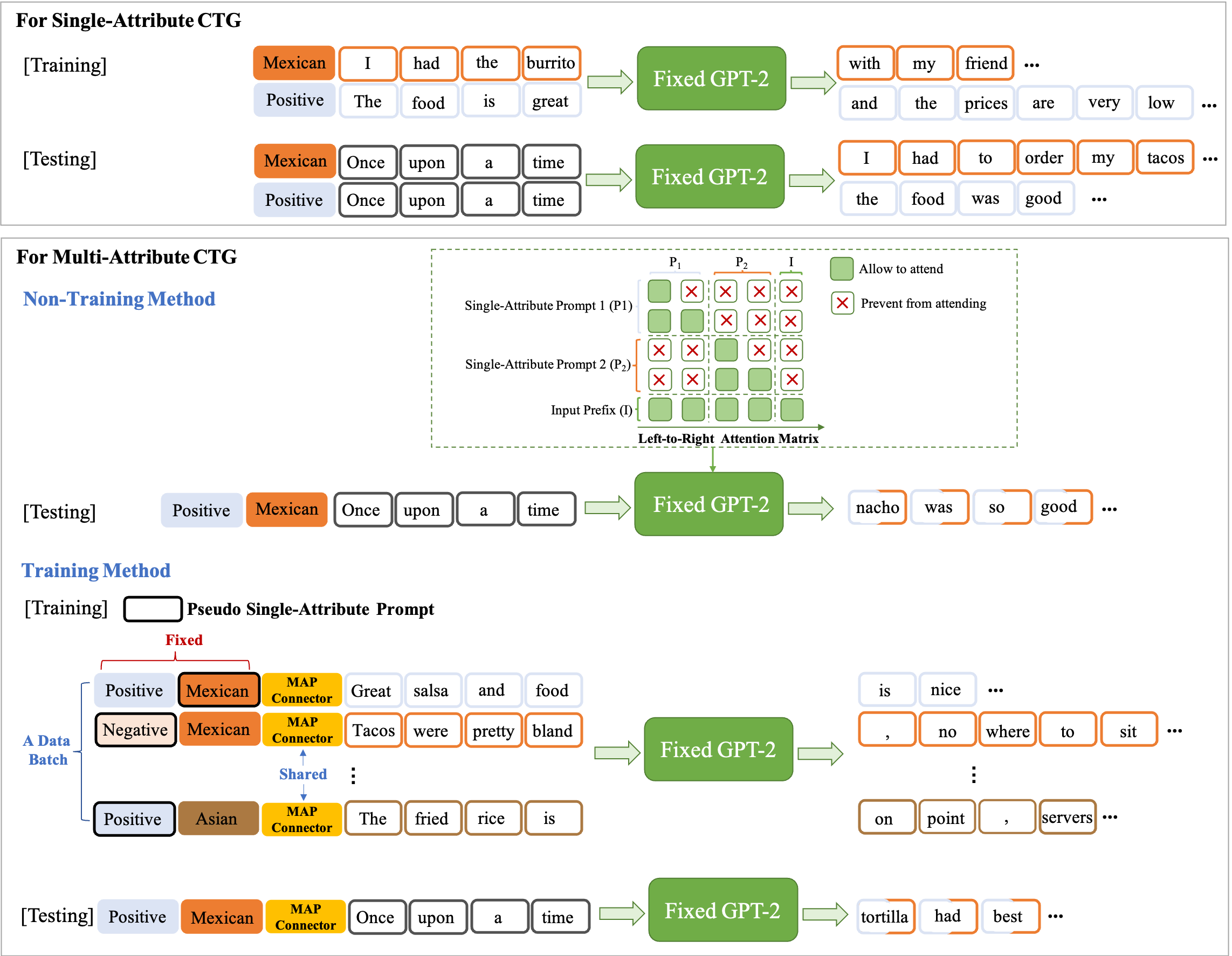}
  \caption{The overview of Tailor to attribute-based CTG. To keep in line with previous works, we ask the fix GPT-2 to continue writing with some attribute-unrelated input prefixes (e.g., Once upon a time) in the testing stage. Notably, the different colored text boxes denote different attribute-specific sentences. For multi-attribute sentences, we use bi-colored text boxes to highlight them.}
  \label{fig:figure2}
\end{figure*}

\section{Methodology}

\subsection{Tailor for Single-Attribute CTG} \label{sec: method_single}
Different from fine-tuning a full copy of PLMs for each attribute, our basic idea is to guide the generation of a PLM with a set of pre-trained continuous vectors, namely single-attribute prompts. Meanwhile, each prompt represents a desirable attribute. As shown in Figure~\ref{fig:figure2} (top), we fix the parameters of a GPT-2 and train each prompt on attribute-specific data. After training, these single-attribute prompts can act as plug-ins for desirable attribute-based text generation. For the conditional prefix ``Once upon a time'', the GPT-2 can continue with ``I had to order my tacos ...'' with a prompt representing the Mexican food topic or `` the food was good'' with a prompt representing the positive sentiment. In this way, our method can be easily expanded: if a new attribute emerges, we only need to train an attribute prompt and then control a PLM to generate attribute-specific sentences.

To be exact, we use language modeling learning object to train such a set of single-attribute prompts. In detail, $k$-th single-attribute prompt $S_{k}$ with length $l_{k}$ is first initialized randomly, where $S_{k} \in \mathbb{R}^{l_{k}\times d_{emb}}$. $d_{emb}$ is the word embedding dimension of the GPT-2. Meanwhile, given an attribute-specific sentence $\bm{x}=\{x_{1},x_{2},...,x_{n}\}$ with length $n$, we get a word sequence matrix $X_{emb}\in \mathbb{R}^{n \times d_{emb}}$ after being embedded by GPT-2. Then, $S_{k}$ is concatenated with $X_{emb}$ to form a input matrix as $[S_{k}; X_{emb}] \in \mathbb{R}^{(l_{k}+n) \times d_{emb}}$, and this matrix is fed into a fixed GPT-2. Finally, the learning object is:

\begin{equation}
\mathcal{L}_{single}=\sum_{t=1}^{n}\log P_{\theta_{g};\theta_{S_{k}}} \left (x_{t} | S_{k},x_{<t} \right),
\end{equation}

\noindent where $\theta_{g}$ and $\theta_{S_{k}}$ denote the parameters of GPT-2 and the single-attribute prompt, respectively. Only $\theta_{S_{k}}$ are updated during the training stage.

\subsection{Tailor for Multi-Attribute CTG} \label{sec: method_multi}
Inspired by composition of discrete prompts~\cite{ptr} to accomplish a complex task, our intuitive idea is to combine single-attribute prompts as a multi-attribute prompt to hint a PLM for multi-attribute CTG. To enjoy the benefit of our paradigm in single attribute CTG, we first consider simply concatenate several single-attribute prompts as a whole multi-attribute prompt. Surprisingly, such a multi-attribute prompt can guide a GPT-2 to generate sentences containing multi attributes of interest, and get encouraging performances in unsupervised settings without any training (see \S~\ref{sec:main_results}). Despite the progress, this straightforward method suffers from notably decreasing fluency of final text compared with single-attribute CTG. Meanwhile, it is position sensitive, i.e., the PLM tends to focus more on the single-attribute prompt that is closer to
the input prefix (see \S~\ref{sec:experiment_discussions}). 

To polish such paradigm while keeping plug-and-play and storage-friendly advantages, as shown in Figure~\ref{fig:figure2} (bottom), Tailor introduces a non-training method to quickly and effectively alleviate the above problems of simply concatenation. Afterward, a training method is further provided to greatly enhance the combinations. We elaborate the two methods separately as follows.

\subsubsection{Non-Training Method}
To make the better use of single-attribute prompts to multi-attribute CTG without any retraining, reducing disparities between the training (a single-attribute prompt for each task) and the testing stage (concatenating more than one single-attribute prompts) is undoubtedly important. Specifically, the single-attribute prompt only attends to itself in the attention matrix while training, as each prompt is individually trained by the attribute-specific data. However, while in the testing stage for multi-attribute CTG, the second prompt also 
focuses on the first one in the concatenation, with the simultaneous change of the position-ids sequence. To fill this gap, MAP mask and RP sequence are introduced to the fixed PLM while generating. MAP mask avoids cross-attention between representations of single-attribute prompts to approximate the condition in the single-attribute CTG training stage. Meanwhile, RP sequence keeps stable prompt position for swapping, preventing such concatenating paradigm from position sensitivity. The details are as follows.

\noindent\textbf{MAP Mask} For the ease of implementation, we introduce MAP mask matrix $M_{p}$ to the softmax logits of GPT-2. Given a vanilla attention module: 
\begin{equation}
    A={\rm Softmax}(\frac{QK^{\top}}{\sqrt{d}}) \in \mathbb{R}^{n\times n} \text{,}
\end{equation}
\noindent where $n$ is the length of input sentence $\bm{x}$ and $Q\text{,}K$ denote representations of query and key, respectively\footnote{The multi-head mechanism is omitted for illustration purposes.}. As to the new attention mask, given two single-attribute prompts $S_{u}$ with length $l_{u}$ and $S_{v}$ with length $l_{v}$, the new attention module is: 

\begin{equation}
\begin{split}
A&={\rm Softmax}(\frac{QK^{\top}}{\sqrt{d}} + M_{p})\in \mathbb{R}^{(l_{p}+n)\times (l_{p}+n)} \text{,} \\
M_{p}^{ij}&=\begin{cases}
    &-\infty ~~i \in [l_{u}, l_{v}]~\text{and}~j \in [0,l_{u}]\text{,}\\
     & 0 ~~~~~~~\text{otherwise,} 
    \end{cases}
\end{split}
\end{equation}

\noindent where $l_{p} = l_{u}+l_{v}$.

\noindent\textbf{RP Sequence} Simply concatenation of single-attribute prompts always suffers from position sensitivity. To address this issue, we propose a simple but effective method to ensure performance consistency while swapping.
In short, we modify the position-ids sequence of the PLM while concatenating. Given the original position-ids sequence:
\begin{equation}
    \bm{id}=\{\underbrace{1, ..., l_{u},}_{\text{Length of $S_{u}$}} \underbrace{l_{u}+1, ..., l_{p},}_{\text{Length of $S_{v}$}} \underbrace{l_{p}+1, ..., l_{p}+n}_{\text{Length of input prefix}}\} \text{,}  
\end{equation}

\noindent the RP sequence can be defined as:
\begin{equation}
    \bm{id_{\text{RP}}}=\{\underbrace{1, ..., l_{u},}_{\text{Length of $S_{u}$}}\underbrace{1, ..., l_{v},}_{\text{Length of $S_{v}$}} \underbrace{l_{v}+1, ..., l_{v}+n}_{\text{Length of input prefix}}\} \text{.}  
\end{equation}

In that case, swapping dose not bring any changes, since the position of prompts is fixed by the RP sequence while avoiding cross-attention by the MAP mask.


\subsubsection{Training Method} \label{sec: method_multi_training}
While the non-training method partly addresses the issues of combination, the inconsistency between the training and testing stage would still decrease the performance. To fill this gap, we provide a training method---MAP connector, which is trained for combining two single-attribute prompts to multi-attribute text generation. To utilize only single-attribute sentences for multi-attribute CTG, we propose a pseudo-attribute prompt based training strategy for MAP connector. 
Therefore, we first detail the pseudo-attribute prompt building method and then the workflow of MAP connector. 

\begin{figure}[h]
  \centering
  \includegraphics[scale=0.39]{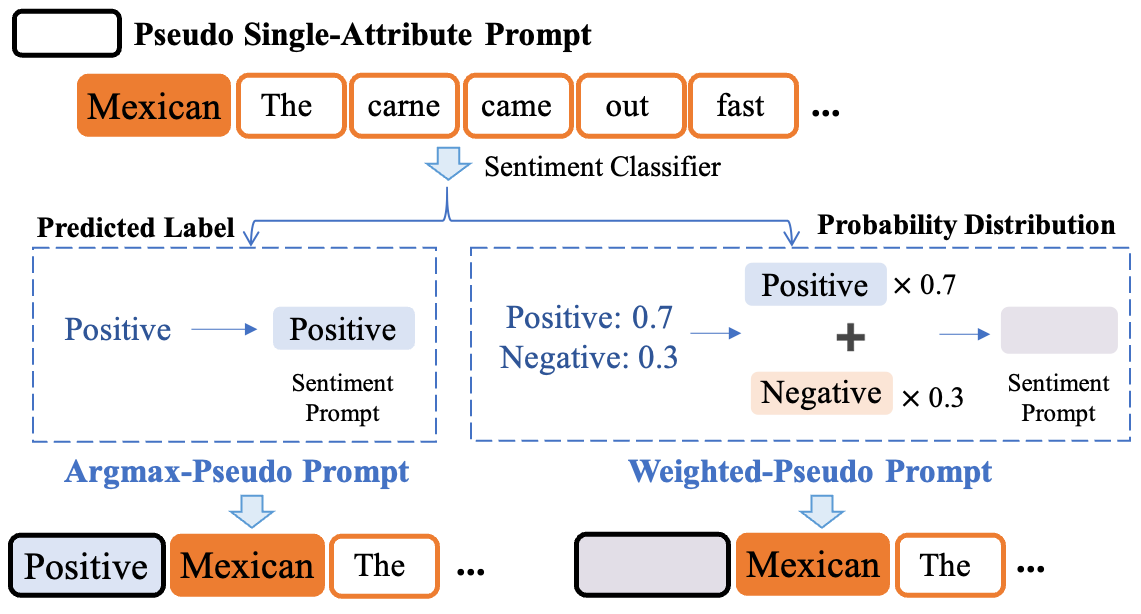}
  \caption{An overview of building persuade-attribute prompt.}
  \label{fig:figure3}
\end{figure}

\noindent\textbf{Building Pseudo Single-Attribute Prompt} Our key idea is to build another pseudo-attribute prompt for each single-attribute sentence, thus MAP connector could be trained in a multi-attribute circumstance. 
An overview of our building method is demonstrated in Figure~\ref{fig:figure3}, where a sentence with the topic of Mexican food is used as a showcase\footnote{In the implementation for multi-attribute CTG, we use YELP data of two sentiments attributes (Positive / Negative) and three kinds of food type (Mexican / American / Asian)}. To be exact, we first train an attribute classifier on the same single-attribute CTG training set. Thus, such a classifier with $n_{class}$ classes corresponds to the pre-trained single-attribute prompt set $\bm{S}=\{S_{1},S_{2}, ..., S_{n_{class}}\}$. Given an attribute-specific sentence $\bm{x}$ of other attribute category, we first get the class probabilities set $\bm{p}=\{p_{1}, p_{2}, ..., p_{n_{class}}\}$. Then, the pseudo single-attribute prompt can be obtained by two methods:

\begin{equation}
\begin{split}
S_{a} &= S_{{\rm Index}\left(\arg\max\left(\bm{p}\right)\right)}\text{,} \\
S_{w} &= \sum_{z=1}^{n_{class}}p_{z}S_{z}\text{,}   
\end{split}
\end{equation}

\noindent where argmax-pseudo prompt method obtains pseudo single-attribute prompt $S_{a}$ by using a single-attribute prompt corresponding to the predicted sentiment, $\textrm{Index} (\cdot)$ means getting the corresponding index. In contrast, weighted-pseudo prompt method utilizes the predicted probability distribution to multiply corresponding single-attribute prompts, respectively. Then these weighted prompt form a whole prompt $S_{w}$ by element-wise addition.

\noindent\textbf{The MAP Connector Workflow} Figure~\ref{fig:figure2} bottom illustrates the workflow of MAP connector. In the training stage, we unify sentences containing different single attributes to train MAP connector, each of which is added an extra pseudo single-attribute prompt (boxes edged in black) by employing the aforementioned method. Specifically, for each training sample, we first concatenate two single-attribute prompts (real and pseudo), MAP connector and the input sentence to a sequence, and then feed it into a fixed GPT-2. It is worth noting that only parameters of MAP connector are updated in the training stage. Therefore, given two single-attribute prompt $S_{u}$ and $S_{v}$, MAP connector $C$ with the length $l_{C}$, $C \in \mathbb{R}^{l_{C}\times d_{emb}}$, we concatenate $S_{u}$, $S_{v}$, $C$ and the input sentence matrix $X_{emb}$ to form a input matrix as $[S_{u};S_{v};C;X_{emb}]$. The learning object is:
\begin{equation}
\mathcal{L}_{multi}=\sum_{t=1}^{n}\log P_{\theta} \left (x_{t} | S_{u}, S_{v}, C, x_{<t} \right),
\end{equation}
where $\theta = [\theta_{g};\theta_{S_{u}};\theta_{S_{v}};\theta_{C}]$. $\theta_{g}$, $\theta_{S_{u}}$, $\theta_{S_{v}}$, and $\theta_{C}$ denote the parameters of GPT-2, two single-attribute prompts and MAP connector, respectively. Only $\theta_{C}$ are updated during the training stage. In the inference stage, we just decompose each multi-attribute generation task as several single-attribute generation tasks and find corresponding single-attribute prompts. Then, these prompts are concatenated with MAP connector to generate sentences that satisfy multi attributes.

\section{Experiments}
\subsection{Experimental Setup} 
\noindent\textbf{Datasets} We conduct experiments on the widely-used benchmark dataset YELP~\cite{multiple_attribute_text} to evaluate Tailor. Following previous works that conduct experiments on attributes of emotions and topics for multi-attribute CTG, we choose Yelp restaurants reviews of sentiment attributes (positive (PO) and negative (NE)) and topics of food type (Mexican (ME), American (AM) and Asian (AS) foods) to evaluate models. Specifically, each attribute contains 30000 / 3000 sentences for training / validation. For evaluation, to keep in line with previous works~\cite{fudge,pplm}, we use 15 attribute-unrelated prefixes\footnote{\url{https://github.com/uber-research/PPLM}} and ask model to continue writing with them (100 sentences for each) while satisfying pre-specified attribute as the final results\footnote{More details can be found in \S~\ref{sec:appendix_dataset}}.
\par 
\noindent\textbf{Evaluation Metrics} Follow \cite{fudge,pplm}, we automatically evaluate generation results from three aspects: (1) \textbf{Correctness}. We used RoBERTa\textsubscript{Large}~\cite{roberta} based attribute classifiers to compute the fraction of final sentences that contains pre-specified attribute, details in Appendix~\ref{sec:appendix_topic_classifiers}. (2) \textbf{Text Quality}. Grammar (GRAM)~\cite{grammar} indicates the averaged grammaticality probabilities of all final sentences, evaluated by a RoBERTa-based CoLA grammaticality model~\cite{fudge}. Perplexity (PPL), we average the scores from GPT-2\textsubscript{Base}, GPT-2\textsubscript{Medium} and GPT-2\textsubscript{Large} version of GPT-2~\cite{gpt2} as the final result. (3) \textbf{Diversity}. Following \citet{diversity}, we report the distinctness of the final results. Specifically, we count the number of unigrams, bigrams and trigrams and then normalize them by the total number of words (i.e., Dist-1 / Dist-2 / Dist-3).
\begin{table*}[h]
\small
\centering
\renewcommand{\arraystretch}{1.2}
\begin{tabular}{llcccc}
\toprule
\multirow{2}{*}{\textbf{TP (\%)}} & \multicolumn{1}{c}{\multirow{2}{*}{\textbf{Method}}} & \multicolumn{1}{c}{\textbf{Correctness}} & \multicolumn{2}{c}{\textbf{Text Quality}} & \multicolumn{1}{c}{\textbf{Diversity}} \\ 
\cmidrule(r){4-5} 
& & (\%) $\uparrow$ & GRAM $\uparrow$ &PPL $\downarrow$ & Dist-1/Dist-2/Dist-3 $\uparrow$\\ 
\midrule
    \multirow{2}{*}{100.00} &FT (Food)   & 87.53  & 0.78 & 40.60 &  0.04 / 0.22  / 0.42  \\
    &FT (Sent) & 97.95 & 0.76 & 42.83 & 0.04 / 0.21  / 0.41  \\
    \multirow{2}{*}{0.10} &Adapter (Food) & 74.70 & 0.75 & 43.85 & 0.04 / 0.23 / 0.46 \\
    &Adapter (Sent) & 93.32 & 0.74 & 47.01 & 0.04 / 0.22 / 0.45  \\
    \multirow{2}{*}{0.001}&PPLM (Food) & 60.64 & 0.34 & 105.33 &  0.16 / 0.53 / 0.80 \\
    &PPLM (Sent)  & 69.37 & 0.36 & 75.59 &  0.15 / 0.53 / 0.82  \\
    \multirow{2}{*}{100.00}&GeDi (Food)  & 99.82 & 0.28 & 278.22 &  0.42 / 0.79 / 0.95  \\
    &GeDi (Sent)  & 87.37  &  0.32 &  517.87 &   0.27 / 0.85 / 0.97  \\
    \cdashline{1-6}[2.5pt/5pt]
    \multirow{2}{*}{0.08}&Tailor-S (Food)  & 83.89 & 0.71 & 45.79 & 0.05 / 0.35 / 0.71 \\
    &Tailor-S (Sent)  & 93.80 & 0.71 & 46.20 & 0.06 / 0.35 / 0.70 \\
\bottomrule
\end{tabular}
\caption{The main results of single-attribute CTG. Due to space constraints, Sent and Food denote averaging the evaluation scores of all sentiment attributes and topics of food type, respectively. TP denotes training parameters as the percentage of the fine-tuning model (FT). $\uparrow$ means higher score is better where $\downarrow$ is exactly the opposite.}
\label{tab:main_single}
\end{table*}
\begin{table*}[h]
\small
\centering
\renewcommand{\arraystretch}{1.25}
\begin{tabular}{llc|ccccc}
\toprule
\multirow{2}{*}{\textbf{TP (\%)}} & \multicolumn{1}{c}{\multirow{2}{*}{\textbf{Method}}} & \multicolumn{3}{c}{\textbf{Correctness (\%)}} & \multicolumn{2}{c}{\textbf{Text Quality}} & \multicolumn{1}{c}{\textbf{Diversity}} \\ 
\cmidrule(r){3-5} \cmidrule(r){6-7} 
& & Avg $\uparrow$ & Sent $\uparrow$ & Food $\uparrow$ & GRAM $\uparrow$ &PPL $\downarrow$ & Dist-1/Dist-2/Dist-3$\uparrow$\\ 
\midrule
100.00 &FT   & 69.80 & 74.03 & 65.57 & 0.69  & 46.54 & 0.04 / 0.23 / 0.42  \\
0.60 &Adapter  & 69.10 & 74.10 & 64.10 & 0.77 & 37.89  &  0.03 / 0.21 / 0.42 \\
 0.60 &Adapter (Pseudo) & 81.71 & 89.95 & 73.46 & 0.75 & 45.63 & 0.04 / 0.22 / 0.45  \\
    \cdashline{1-8}[2.5pt/5pt]
0.00 & CONCAT    &  76.20 &  87.88 & 64.51 & 0.63 & 55.02 & 0.05 / 0.33 / 0.68  \\
0.00 &Tailor-C    & 78.82 &  87.54 & 70.10 & 0.63 & 52.76 & 0.05 / 0.32 / 0.68  \\
0.08 &Tailor-W     & 83.98  & 93.27 & 74.68 & 0.68 & 51.41 & 0.05 / 0.33 / 0.69 \\
 0.08 &Tailor-A    & 87.15 & 92.97  & 81.32 & 0.69 & 52.73 & 0.05 / 0.33 / 0.69 \\
\bottomrule
\end{tabular}
\caption{The main results of multi-attribute CTG. Due to space constraints, we average the scores of six combinations (two sentiment attributes $\times$ three topic attributes of food type) as the final results for each method.}
\label{tab:main_multi}
\end{table*}

\noindent\textbf{Tailor Settings} To facilitate description, for single-attribute CTG, Tailor-S denotes the method of single-attribute prompts. While for multi-attribute CTG, CONCAT means simply concatenating two single-attribute prompts and Tailor-C is our non-training method. For the training method, Tailor-A and Tailor-W represent using argmax-pseudo and weighted-pseudo prompts in the training stage of the MAP connector, respectively.

\noindent\textbf{Baselines} We compare our methods with mainstream competitive models as follows. (1) FT, fine-tuning the original GPT-2\textsubscript{Base} on attribute-specific data. As multi-attribute CTG is unsupervised, following ~\citet{style_ptb}, we sequentially apply the GPT-2 trained for corresponding single-attribute data multiple times to perform multi-attribute CTG. (2) Adapter, following~\citet{prefix-tuning},  we use the adapter for GPT-2 as same as \citet{vglm}. Note that for multi-attribute CTG, we first use the same training method as mentioned in FT for Adapter. Besides, we use the same argmax-pseudo labeled sentences (see \S~\ref{sec: method_multi_training}) to train the Adapter (marked with `Pseudo').
(3) GeDi~\cite{gedi}, using small PLMs to hint large ones. (4) PPLM~\cite{pplm}, back-propagating gradients of extra attribute classifiers to a PLM. Due to space constraints, the implementation details of baselines and Tailor can be found in \S~\ref{sec:appendix_impletemt}.

\subsection{Main Results} \label{sec:main_results}
\noindent \textbf{Single-Attribute CTG} As shown in Table~\ref{tab:main_single}, Tailor-S outperforms PPLM and GeDi to a great extend on both correctness and text quality. Meanwhile, compared with other parameter-efficient learning model Adapter, Tailor-S also gets improvements on both on both correctness (e.g, + 9.19\% of Food) and diversity (e.g, + 0.02\% / + 0.12\% / + 0.25\% of Food) with a similar scale of training parameters. However, with 0.08\% training parameters of the GPT-2, Tailor-S still has a performance gap with FT, e.g., - 4.14\% correctness on Food. Fortunately, as the length of Tailor-S increases (see \S~\ref{sec:experiment_discussions}), this gap appears to narrow (- 0.33\%, Tailor-S with length of 256).

\noindent \textbf{Multi-Attribute CTG} As shown in Table~\ref{tab:main_multi}, we compare three instantiations of Tailor and strong baselines in the single-attribute CTG experiment. First, Tailor-C shows encouraging performances without any training, especially on correctness, outperforms fine-tuning (+ 13.51\% Sentiment / + 4.53\% Food) and Adapter (+ 13.44\% Sentiment / + 6.00\% Food), yet text quality decrease. Besides, our training methods Tailor-W and Tailor-A show improvements on all scores compared with Tailor-C, e.g., +  4.58\% / + 11.22\% correctness on the topic of food type attribute. Meanwhile, Tailor also outperforms Adapter with the same pseudo label strategy on both correctness and diversity, with a notable scale discrepancy of training parameters (1:7.25).

\begin{table}[h]
\small
\centering
\renewcommand{\arraystretch}{1.2}
\begin{tabular}{lll|ll}
\toprule
\multirow{2}{*}{\textbf{TP (\%)}} &\multirow{2}{*}{\textbf{Method}} & \multicolumn{3}{c}{\textbf{Correctness (\%)}} \\ 
\cmidrule(r){3-5} 
& & Avg$\uparrow$& Sent$\uparrow$ & Food$\uparrow$\\ 
\midrule
    \multicolumn{5}{c}{Single-Attribute CTG} \\
    \multirow{2}{*}{100.00} &FT & 54.08 &  - &  54.08 \\
    &FT &85.28 & 85.28 & - \\
    \multirow{2}{*}{0.10} &Adapter &55.79 & - & 55.79 \\
    &Adapter & 77.91 & 77.91 & -  \\
    \cdashline{1-5}[2.5pt/5pt]
    \multirow{2}{*}{0.08}&Tailor-S &66.23 & - & 66.23 \\
    &Tailor-S  & 89.27 & 89.27 & - \\
\midrule
    \multicolumn{5}{c}{Multi-Attribute CTG} \\
100.00&FT   & 60.60 & 73.45 & 47.75 \\
0.60 &Adapter  & 57.15 & 68.44 & 45.85 \\
0.60 &Adapter (Pseudo)    & 67.27 & 78.66 & 55.88 \\
    \cdashline{1-5}[2.5pt/5pt]
0.00 & Tailor-C & 68.09 & 74.38 & 61.79 \\
0.08 & Tailor-W & 70.32 & 84.18 & 56.46 \\
0.08 & Tailor-A & 71.41 & 83.63 & 59.18  \\
\bottomrule
\end{tabular}
\caption{The main results of few-shot learning. Note that TP for multi-Attribute CTG means the extra training parameters as the percentage of the fine-tuning model (FT) after single-Attribute CTG. Due to space constraints, we average the scores of six combinations (two sentiment attributes $\times$ three topic attributes of food type) as the final results for each method.}
\label{tab:main_few_shot_short}
\end{table}
\subsection{Further Discussions} \label{sec:experiment_discussions}
\noindent\textbf{Few-Shot Learning} We conduct a few-shot learning setting to further analyze the effectiveness of Tailor. In detail, following \citet{prefix-tuning}, we randomly sample from full dataset and obtain the few-shot dataset (training / validation / testing: 150 / 20 / 20). Specifically, we sample three different few-shot datasets and average the scores of each method on three datasets as the final results. As shown in Table~\ref{tab:main_few_shot_short}, three types of Tailor outperforms other baselines on correctness, with 0.00\% / 0.08\% extra training parameters of the GPT-2 after single-attribute CTG. 

\begin{figure}[t]
  \centering
  \includegraphics[scale=0.37]{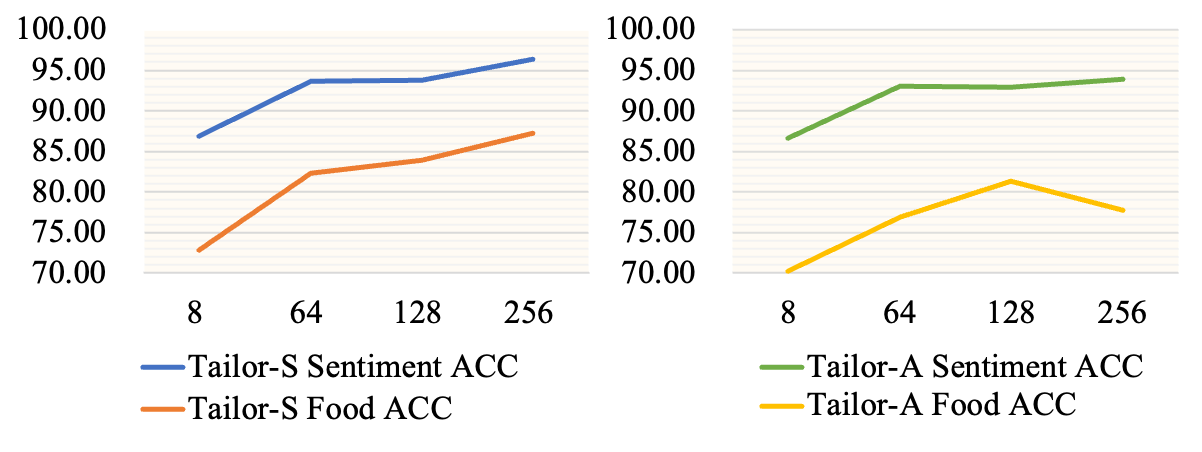}
  \caption{The results of using Tailor-S and Tailor-A with different lengths. The x-axis is the prompt length and the y-axis is the averaging correctness score (\%).}
  \label{fig:length}
\end{figure}

\noindent\textbf {Length of Tailor} As shown in Figure~\ref{fig:length}, we explore the length of both Tailor-S and Tailor-A. For singe-attribute prompt Tailor-S, the performances increase alongside the length. But for Tailor-A, it obtains the best performances with the length of 128, and the performances have a slight drop when we continue to increase the length. \par
\noindent\textbf{Position Sensitivity} We investigate the position sensitivity problem when concatenating two single-attribute prompts. As shown in Table~\ref{tab:main_sensitivity}, for simply concatenation, the GPT-2 tends to focus more on the prompt that is closer to the input prefix (i.e., the attribute behind the dash in the Table~\ref{tab:main_sensitivity}). For instance, NE attribute get a 3.14\% improvement if we put the corresponding prompt close to the input prefix. However, it also brings a 3.4\% decrease for AM attribute as being away from input prefix at the same time. In contrast, Tailor-C keeps the same performance after swapping. 

\begin{table}[h]
\small
\centering
\renewcommand{\arraystretch}{1.2}
\begin{tabular}{lll|lll}
\toprule
\multirow{2}{*}{\textbf{Method}}& \multirow{2}{*}{\textbf{Combination}} & \multicolumn{3}{c}{\textbf{Correctness (\%)}} \\
\cmidrule(r){3-5}
& & Avg$\uparrow$ & Sent$\uparrow$ & Food$\uparrow$\\ 
\midrule
\multirow{2}{*}{CONCAT} & NE-AM & 68.40 & 76.93 & 59.87 \\
& AM-NE & 68.27 & 80.07 & 56.47 \\
\cdashline{1-5}[2.5pt/5pt]
\multirow{2}{*}{Tailor-C} &NE-AM & 69.90 & 79.07 & 60.73 \\
 &AM-NE & 69.90 & 79.07 & 60.73 \\
\bottomrule
\end{tabular}
\caption{The results on multi-attribute CTG of generating sentences satisfying negative
sentiment (NE) and topic of American food (AM). NE-AM denotes putting the positive attribute prompt in first and American food attribute prompt in later when concatenating them, in contrast to AM-NE.}
\label{tab:main_sensitivity}
\end{table}

\noindent\textbf{Ablation Study of Tailor-C} Whether Tailor-C enjoys the benefits from the MAP mask and the RP sequence is also of concern. As shown in Table~\ref{tab:main_attention}, both the MAP mask and the RP sequence are important to Tailor-C. More importantly, they are complementary to each other---using these two strategies simultaneously can improve the
performance while avoiding the position sensitivity problem.\par 
\begin{table}[h]
\small
\centering
\renewcommand{\arraystretch}{1.2}
\begin{tabular}{ll|lll}
\toprule
\multirow{2}{*}{\textbf{Method}}& \multicolumn{3}{c}{\textbf{Correctness (\%)}} \\
\cmidrule(r){2-4}
& Avg$\uparrow$ & Sent$\uparrow$ & Food$\uparrow$\\ 
\midrule
Tailor-C & 78.82 & 87.54 & 70.10 \\
- APA Mask & 78.36 & 87.39 & 69.34 \\
- RP & 77.77  & 88.33 & 67.21 \\
- Both & 76.20 & 87.88 & 64.52 \\
\bottomrule
\end{tabular}
\caption{The ablation study on using the APA mask and the RP sequence (RP) of Tailor-C. `-' denotes removing the corresponding module from Tailor-C. Note that, exchanging the concatenating order of prompts would bring different performances, except for Tailor-C. Thus, we average the scores from these two situations of six attributes combinations as the final result.}

\label{tab:main_attention}
\end{table}

\noindent\textbf{Unseen Combination} In this part, we analyze the combining ability of Tailor on the unseen combination, which does not appear in Tailor's training stage. In implementation, we randomly select one combination and remove the corresponding data from the training set for the MAP connector, and then test the performance of the MAP connector on this multi-attribute generation task. As shown in Table~\ref{tab:main_unseen}, Tailor-A still works to the unseen combination PO-ME, and outperforms the non-training method Tailor-C with 2.35\% improvements.\par
\begin{table}[h]
\small
\centering
\renewcommand{\arraystretch}{1.2}
\begin{tabular}{lll|lll}
\toprule
\multirow{2}{*}{\textbf{Unseen}}& \multirow{2}{*}{\textbf{Method}} & \multicolumn{3}{c}{\textbf{Correctness (\%)}} \\ 
\cmidrule(r){3-5}
& &Avg$\uparrow$ & Sent$\uparrow$ & Food$\uparrow$\\ 
\midrule
PO-ME & Tailor-C & 87.54 & 95.60 & 79.47 \\
PO-ME & Tailor-A & 89.89 & 97.07 & 82.70 \\
-& Tailor-A & 91.64 & 97.87 & 85.40 \\
\bottomrule
\end{tabular}
\caption{The results on unseen combination to multi-attribute CTG. PO-ME denotes the attribute combination of positive sentiment and topic of Mexican food. }
\label{tab:main_unseen}
\end{table}

\section{Conclusions}
In this paper, we explore attribute-based CTG in a prompt-based manner---Tailor, which represents each attribute as a continuous prompt and effectively combines them as a multi-attribute prompt. For enhancing these combinations, Tailor provides two solutions, namely non-training (MAP mask + RP sequence) and training methods (MAP connector). As our first attempt to multi-attribute CTG, combining more than two attributes 
still needs to be discussed. Thus in the future, we will investigate extending Tailor to connect wider ranges of attributes, and expand it to other text-to-text generation tasks.      
\bibliography{refer}
\bibliographystyle{acl_natbib}

\appendix
\section{Implement Details} \label{sec:appendix_impletemt}
We detail the hyperparameters and experimental settings of Tailor and baselines as follows.

\begin{enumerate}[leftmargin=*]
    \item Tailor. Tailor is implemented based on Huggingface~\cite{huggingface}. In all experiments of Tailor, we set the length of Tailor-C to 128, as same as the MAP connector for Tailor-A and Tailor-W. As for the learning rate and the warm-up steps, Tailor-S, Tailor-A, and Tailor-W are set to 5e-5 and 0, respectively. Besides, to get a pseudo label for MAP connector, we use the RoBERTa\textsubscript{Large} based classifier for both sentiment and topic of food type attributes. The hyperparameters can be found in \S~\ref{sec:appendix_topic_classifiers}. Note that, for a fair comparison, we only use the same training set for each classifier as for training Tailor.
    \item FT\footnote{\url{https://huggingface.co/gpt2}}. We use the GPT-2\textsubscript{Base} with a language model head implemented based on Huggingface. The learning rate is set to 5e-3 and the warm-up steps is set to 0.
    \item Adapter\footnote{\url{https://github.com/zlinao/VGLM}}. we set the bottleneck size to 5 to keep a similar size of training parameters with Tailor. The learning rate is set to 5e-5 and the warm-up steps is set to 0.
    \item GeDi\footnote{\url{https://github.com/salesforce/GeDi}}. For a fair comparison, we use the generative discriminator of GeDi based on GPT-2\textsubscript{Base} to guide generation of another GPT-2\textsubscript{Base}. In inference, we use the $\omega=30$, $\rho=0.8$ and $\tau=0.8$, as reported in their implementation.
    \item PPLM\footnote{\url{https://github.com/uber-research/PPLM/blob/master/paper_code/pplm.py}}. We employ the original hyper-parameter setting reported in \citet{pplm}. In detail, $\gamma=1.5$, $\gamma_{gm}=0.9$, $\lambda_{kl}=0.01$, iterations=3 and step size=0.02.
\end{enumerate}


In inference, to keep in line with previous works~\cite{pplm,gedi}, we use top-$k$ sampling with $k$=10, and fix the random seed as 42 for all models to get the final results, while the maximum generation length is set to 128.

\section{Yelp Dataset} \label{sec:appendix_dataset}
In this section, we elaborate the workflow of filtering, pre-processing and sub-sampling to get the attribute-specific dataset for training all models and the classifiers For correctness evaluation. First of all, we get the YELP dataset from \citet{multiple_attribute_text}. In detail, each sample of the YELP dataset contains a review and the corresponding attributes\footnote{The format can be found via \url{https://github.com/shrimai/Style-Transfer-Through-Back-Translation}}. Then, we select the restaurant reviews sub set as our original dataset. For dataset filtering, we use the dataset setup scripts offered by \citet{multiple_attribute_text}, which contains a fastText\cite{fasttext} classifier to filter sentences that not written in English. After that, we filter the sentences with rated 3 stars, since they could be neutral in sentiment~\cite{filter_stars}. Finally, we get the pre-processed dataset as illustrated in Table~\ref{tab:appendix_dataset}. For the classifiers that are used in correctness evaluation, we use the full dataset and details in \S~\ref{sec:appendix_topic_classifiers}. Aside from it, for training Tailor and baselines, we randomly sample 30,000 / 3,000 sentences as training / validation data set for each attribute.
\begin{table}[h]
\centering
\begin{tabular}{lc}
\toprule
Model                & F1 Score \\
\midrule
Food Type Classifier & 83.40    \\
Sentiment Classifier & 97.10   \\
\bottomrule
\end{tabular}
\caption{The Performances of two classifiers on Yelp dataset.}
\label{tab:appendix_classifier}
\end{table}

\begin{table}[h]
\centering
\renewcommand{\arraystretch}{1.2}
\begin{tabular}{lccc}
\toprule
Attribute & PO & NE & All    \\ \midrule
ME    & 25,169              & 89,411              & 114,580 \\
AM  & 72,641              & 299,293             & 371,934 \\
AS     & 47,680              & 185,551             & 233,231 \\
All            & 145,490             & 574,255             & 719,745 \\ \bottomrule
\end{tabular}
\caption{The number of reviews for each attribute in Yelp dataset.}
\label{tab:appendix_dataset}
\end{table}

\section{Classifiers For Correctness Evaluation} \label{sec:appendix_topic_classifiers}
We use the RoBERTa\textsubscript{Large} based model to train two classifiers for both sentiment and topic of food type attributes. To obtain a balanced dataset, we randomly over-sampling the raw dataset. Finally, we get 1500k / 15k / 15k topic-specific sentences and 1380k / 1k / 1k sentiment-specific sentences for training / validation / testing, respectively. For training two classifiers, the learning rate is set to 5e-5 and the warm-up steps is set to 200. The performances on the testing set can be found in Table~\ref{tab:appendix_classifier}.


\section{Case Study}
To intuitively display the effects of various attributes, we show some generation results of single-attribute CTG in Table~\ref{tab:appendix_case_single} and multi-attribute CTG in Table~\ref{tab:appendix_case_multi}, respectively.

\begin{table*}[h]
\centering
\begin{tabular}{ccp{12cm}}
\toprule
\textbf{Attribute} & \textbf{Method} & \textbf{Generation Results} \\
\multirow{10}{*}{\colorbox{c1}{NE}}     &   FT   & Once upon a time, i was very \colorbox{c1}{disappointed}. The meat was \colorbox{c1}{bland} and the beans tasted as if they had been sitting out all day...    \\
                        &    Adapter    & Once upon a time in the restaurant it was \colorbox{c1}{still dark} and people weren 't even talking... \\ 
                        &    PPLM   & Once upon a time, computers would have been able read, interpret and write, and listen, listen and read...                   \\ 
                        &    GeDi   & Once upon a time you either enter base build states or begin switching context switches and magic spells that alter your manifest...                    \\ 
                        &    Tailor-S   & Once upon a time, you had to order your dinner. the food \colorbox{c1}{came out cold} with \textbf{no} seasoning or flavor whatsoever... \\ 
\midrule
\multirow{10}{*}{\colorbox{c2}{AS}} &   FT   & Once upon a time I've had the \colorbox{c2}{spicy tofu dish}, but that was my only meal. It came out cold and tasted awful...                    \\
                        &    Adapter    & Once upon a time i was craving something spicy, it tasted like the best \colorbox{c2}{Chinese food} out there...                    \\ 
                        &    PPLM   & Once upon a time I made a stone of silver ring mail "Garden of the Winds Winds"... \\ 
                        &    GeDi   & Once upon a time bamboo \colorbox{c2}{noodles} were the classical medicine and lemongrass fetish... \\ 
                        &    Tailor-S   & Once upon a time, I got here for the \colorbox{c2}{sushi roll}. After getting home from work at 4pm and finding... \\ 
\bottomrule
\end{tabular}
\caption{Samples of single-attribute CTG with input prefix `Once upon a time'. NE and AS denotes generating sentences satisfying negative sentiment and topic of Asian food, respectively. We highlight different attribute-specific words or phrases for better view.}
\label{tab:appendix_case_single}
\end{table*}

\begin{table*}[h]
\centering
\begin{tabular}{ccp{11cm}}
\toprule
\textbf{Attribute} & \textbf{Method} & \textbf{Generation Results} \\
\multirow{12}{*}{\colorbox{c1}{NE}-\colorbox{c2}{AS}} &   FT   & Once upon a time I was greeted, sat and waited patiently. the food \colorbox{c1}{took forever} and there were only 6 of us that got our appetizers...    \\
                        &    Adapter    &Once upon a time I got my food and was told that the service is \colorbox{c1}{slow}. then they came over to me with an \colorbox{c1}{"error"}... \\ 
                        &    Adapter (P) &Once upon a time, I would \colorbox{c1}{never recommend} eating this place. the \colorbox{c2}{sushi} was \colorbox{c1}{terrible} and they... \\ 
                        &    Tailor-C   & Once upon a time, my mom had to order the \colorbox{c2}{fried rice} at night and she said that it was \colorbox{c1}{so bad}... \\ 
                        &    Tailor-A   & Once upon a time, I've had my \colorbox{c2}{rice and noodles} at the \colorbox{c2}{Japanese buffet}. They were \colorbox{c1}{so bland} that... \\ 
                        &    Tailor-W   & Once upon a time I had \colorbox{c2}{the spicy ramen}. It was \colorbox{c1}{too sweet and salty}, but now its like they have been replaced with something else... \\ 
\bottomrule
\end{tabular}
\caption{Samples of multi-attribute CTG with input prefix `Once upon a time'. NE-AS denotes generating sentences satisfying negative sentiment and topic of Asian food. Adapter (P) denotes using the same argmax-pseudo labeled sentences (see \S~\ref{sec: method_multi_training}) to train the Adapter. We highlight different attribute-specific words or phrases for better view.}
\label{tab:appendix_case_multi}
\end{table*}

\end{document}